\documentclass{article}
\usepackage[final]{nips_2018}

\usepackage[utf8]{inputenc} 
\usepackage[T1]{fontenc}    
\usepackage{hyperref}       
\usepackage{url}            
\usepackage{booktabs}       
\usepackage{amsfonts}       
\usepackage{nicefrac}       
\usepackage{microtype}      
\usepackage{comment}
\usepackage{times}
\usepackage{latexsym}
\usepackage{graphicx}
\usepackage{authblk}
\usepackage{hyperref}

\usepackage{url}

\title{Sequence Learning with RNNs for Medical Concept Normalization in User-Generated Texts}

\author[1,4]{Elena Tutubalina}
\author[1,4]{Zulfat Miftahutdinov}
\author[1,2,4]{Sergey Nikolenko}
\author[3]{Valentin Malykh}
\affil[1]{\textit{Chemoinformatics and Molecular Modeling Laboratory, Kazan Federal University, 18 Kremlyovskaya street, Kazan, Russian Federation, 420008}}
\affil[2]{\textit{Laboratory of Mathematical Logic, St. Petersburg Department of the Steklov Mathematical Institute, 27 Fontanka, St.Petersburg, Russian Federation, 191023}}
\affil[3]{\textit{Moscow Institute of Physics and Technology, Dolgoprudny, Russia, 141700}}
\affil[4]{\textit{Neuromation OU, Tallinn, 10111 Estonia}}

\date{}

\begin{document}
\maketitle
\begin{abstract}
In this work, we consider the \emph{medical concept normalization} problem, i.e., the problem of mapping a disease mention in free-form text to a concept in a controlled vocabulary, usually to the standard thesaurus in the Unified Medical Language System (UMLS).
This task is challenging since medical terminology is very different when coming from health care professionals or from the general public in the form of social media texts. We approach it as a sequence learning problem, with recurrent neural networks trained to obtain semantic representations of one- and multi-word expressions. We develop end-to-end neural architectures tailored specifically to medical concept normalization, including bidirectional LSTM and GRU with an attention mechanism and additional semantic similarity features based on UMLS. Our evaluation over a standard benchmark shows that our model improves over a state of the art baseline for classification based on CNNs. 
\end{abstract}

\section{Introduction}
User-generated texts (UGT) on social media present a wide variety of facts, experiences, and opinions on numerous topics, and this treasure trove of information is currently severely underexplored. We consider the problem of discovering medical concepts in UGTs with the ultimate goal of mining new symptoms, adverse drug effects, and other information about a disease or a drug. 
An important part of this problem is to translate a text from ``social media language'' (e.g., ``can't fall asleep all night'' or ``head spinning a little'') to ``formal medical language'' (e.g., ``insomnia'' and ``dizziness'' respectively). This is necessary to match user-generated descriptions with medical concepts, but it is more than just a simple matching of UGTs against a vocabulary.
We call the task of mapping the language of UGTs to medical terminology \textit{medical concept normalization}. 
It is especially difficult since UGTs on social media patients discuss different concepts of illness and a wide array of drug reactions. Moreover, UGTs from social networks are very noisy, containing misspelled words, incorrect grammar, hashtags, abbreviations, smileys, different variations of the same word, and so on. 


Some recent works have tried to view the problem of matching a one- or multi-word expression against a knowledge base as a supervised sequence labeling problem. \cite{Han201749} and \cite{Belousov201754} utilize recurrent neural networks (RNNs) for phrase normalization in tweets and provide similar performances, while~\citet{limsopatham2016normalising} applied convolutional neural networks (CNNs) to user reviews. These works were among the first applications of deep learning methods for medical concept normalization. 
In this work, we extend previous research with state of the art RNN architectures with an attention mechanism and special provisions for domain knowledge. The artitectures utilize Long Short-Term Memory (LSTM) units~\citep{lstm97and95} and Gated Recurrent Units (GRU)~\citep{cho2014learning}. We have conducted experiments on a real-life dataset from \emph{Askapatient.com}, the CADEC corpus~\citep{karimi2015cadec}, proving the effectiveness of our approach. This work is an abridged version of our recent paper~\citep{tutubalina2018medical}.

\section{Methods}
We used bidirectional LSTM and GRU units with an attention mechanism on top of the embedding layer, with hyperbolic tangent ($\tanh$) activations. The output from the RNN layer is concatenated with a number of semantic similarity features based on prior knowledge. Lastly, we add a softmax layer to convert values to conditional probabilities. We adopted two kinds of word embeddings: (i) \emph{HealthVec} trained on $2{,}5$M health-related reviews~\citep{miftahutdinov2017} and (ii) \emph{PubMedVec} trained on biomedical literature~\citep{pyysalo2013distributional}.


\subsection{Semantic Similarity Features}
We have extracted a set of features to enhance the representation of phrases. These features contain cosine similarities between the vectors of an input phrase and a concept in a medical terminology dictionary. The dictionary includes medical codes and synonyms from the \emph{UMLS Metathesaurus} (version 2017 AA), where codes are presented in the CADEC corpus. We have applied and compared three strategies to constructing representations of a concept and a mention based on cosine similarity between the representations of each pair:
\begin{itemize}
\item \textsc{TF-IDF (all)}: represent a medical code as a single document by concatenating synonymous terms; then, apply the TF-IDF transformation on the code and the entity mention and compute the cosine similarity;
\item \textsc{TF-IDF (max)}: represent a medical code as a set of terms; for each term, compute the cosine distance between its TF-IDF representation and the entity mention and then select the largest similarity;
\item \textsc{w2v (all)}: represent a medical code as a single document by concatenating synonymous terms; then, embed a code and an entity mention as averaged sums of the embeddings of its words and compute the cosine similarity.
\end{itemize}

\section{Evaluation}
\subsection{Datasets} 
We have conducted our experiments on a collection of $1{,}250$ user reviews (UGTs) obtained from the CADEC corpus~\citep{karimi2015cadec}. Only $39.4$\% annotations (including drugs) were unique; people generally discussed similar reactions. All entities in the CADEC corpus were mapped to SNOMED CT-AU by a clinical terminologist. We removed ``conceptless'' or ambiguous mentions for the purposes of evaluation. There were $6{,}754$ entities and $1{,}029$ unique codes in total.

\subsection{Experiments} 
We have performed 5-fold cross-validation to evaluate the methods, discovering that a standard cross-validation method creates a high overlap of expressions in exact matching between training and testing parts (approx.~$40$\%). Therefore, the train/test splitting procedure has a special form in our setup: we removed duplicates of mentions and grouped medical records we're working with into sets related to specific medical codes. Then, each set has been split independently into $k$ folds, and all folds have been merged into the final $k$ folds named \textit{DisEntFolds}. 
For comparison, we have applied state-of-the-art architectures based on CNNs. We adopted effective parameters from~\cite{kim2014convolutional} and \cite{limsopatham2016normalising}, using a filter $w$ with window size $h\in\{3, 4, 5\}$, each with $100$ feature maps. Pooled features are fed to a fully connected feed-forward neural network (with dimension $100$) to perform inference, using rectified linear units (ReLU) as the output activation.

We present the experimental results of our neural networks in Table~\ref{tab:results}. Models with semantic similarity features are labeled with~\textsc{TF-IDF}, with similarities based either on all synonymous terms (``all'') or the term with the largest similarity (``max''). Attention-based GRU with prior knowledge  achieved the accuracy of $70.05$\%. The best results have been obtained with vectors trained on social media posts. Table~\ref{tab:results} shows that the attention mechanism and additional features indeed lead to quality improvements for both GRU and LSTM.  We also compare our model and existing state-of-the-art results on \emph{AskAPatient} folds generated from the CADEC corpus in Table~\ref{tab:compare}.

\begin{table}[t!]
\caption{The accuracy performance on DisEntFolds.}\label{tab:results}
\centering
\scalebox{0.85}{
\begin{tabular}{|l|l|c|}
\hline
\textbf{Model} & \textbf{Parameters} & \textbf{Acc.} \\ \hline
CNN & HealthVec, 100 feature maps & 46.19 \\ 
CNN & PubMedVec, 100 feature maps & 45.79 \\ \hline
LSTM & HealthVec, 200 hidden units & 64.51 \\ 
LSTM & PubMedVec, 200 hidden units &64.24 \\ 
GRU & HealthVec, 200 hidden units &  63.05\\ 
GRU & PubMedVec, 200 hidden units & 62.73 \\ \hline
LSTM+A. & HealthVec, 200 hidden units & 65.73\\ 
GRU+A. & HealthVec, 200 hidden units & \textbf{67.08} \\ 
GRU+A. & PubMedVec, 200 hidden units & 66.55 \\ 
GR+A. & HealthVec, 100 hidden units & 66.56 \\ 
\hline
LSTM+A. & HealthVec, 100, \textsc{TF-IDF (all)} & 67.63\\ 
LSTM+A. & HealthVec, 200, \textsc{TF-IDF (all)} & 66.83 \\ 
GRU+A.& HealthVec, 100, \textsc{TF-IDF (all)} & 69.92 \\ 
GRU+A. & HealthVec, 200, \textsc{TF-IDF (all)} & 69.42 \\ 
GRU+A. & HealthVec, 100, \textsc{w2v sim. (all)} & 69.14 \\ 
GRU+A. & HealthVec, 100, \textsc{TF-IDF (max)}& \textbf{70.05}\\ 
\hline
\end{tabular}
}
\end{table}

\begin{table}[]
\centering
\caption{The accuracy performance on the AskAPatient folds from \cite{limsopatham2016normalising}.}
\label{tab:compare}
\scalebox{0.8}{
\begin{tabular}{|l|c|}
\hline
\multicolumn{1}{|l|}{\textbf{Model}} & \multicolumn{1}{l|}{\textbf{Accuracy}} \\ \hline
DNorm \cite{limsopatham2016normalising} & 73.39 \\
CNN \cite{limsopatham2016normalising} & 81.41 \\ 
RNN \cite{limsopatham2016normalising} & 79.98 \\  \hline
Multi-Task Attentional Character-level CNN \cite{niu2018multi} & 84.65 \\  \hline
GRU+Attention (HealthVec, \textsc{TF-IDF (max)}) & 85.71 \\ \hline
\end{tabular}
}\vspace{.5cm}

\parbox{.48\linewidth}{
\centering
\caption{Summary statistics of entity mention phrases of different length.}
\label{tab:phsum}
\begin{tabular}{|c|c|}
\hline
\multicolumn{1}{|l|}{\textbf{Length of a mention}} & \multicolumn{1}{l|}{\textbf{\# mentions}} \\ \hline
1 & 11 \\ 
2 & 558 \\ 
3 & 1064 \\ 
4 & 739 \\ 
5 & 531 \\ 
6 or longer & 662 \\ \hline
\end{tabular}
}
\hfill
\parbox{.48\linewidth}{
\caption{The performance of GRU+Attention with similarity features \textsc{TF-IDF (max)}.}
\label{tab:phacc}
\begin{tabular}{|c|c|}
\hline
\multicolumn{1}{|l|}{\textbf{Length of a mention}} & \multicolumn{1}{l|}{\textbf{Accuracy}} \\ \hline
1 & 67.16 \\ 
2 & 71.43 \\ 
3 & 72.91 \\ 
4 & 72.15 \\ 
5 & 65.49 \\ 
6 or longer & 50.12 \\ \hline
\end{tabular}
}
\end{table}

We have evaluated the performance of our best model on entity mention phrases of different lengths from \emph{DisEntFolds}. As can be seen from the results in Table~\ref{tab:phacc}, GRU with attention and similarity features \textsc{TF-IDF (max)} achieved the best results at $72.91$\% accuracy and $72.15$\% accuracy on three-word and four-word expressions respectively. 

\subsection{Error Analysis}
The primary goal of our model was to map disease-related mentions into medical codes. We evaluated the capability of GRU-based architectures with attention in fulfilling this goal and examined the output of our network. One important limitation of deep learning technique is the need for sufficient training data; RNNs do not perform well on rare, long, discontiguous, and overlapping expressions. For example, the mention ``toes became so painful'' is automatically associated with the concept ``Pain'' (SNOMED ID 22253000), while the ground truth label is the less frequent concept ``Pain in toe'' (SNOMED ID 285365001). Handling discontiguous expressions may involve word reordering: ``toes became so painful'' into ``painful toes''. Another group of errors is related to expressions that overlap in meaning and share similar characteristics. For example, the model associates the mention ``could only walk less than 100 meters'' with the concept ``Walking disability'' (SNOMED ID 228158008), while the ground truth concept is ``Reduced mobility'' (SNOMED ID 8510008). The mention ``foggy thinking'' is automatically associated with ``Unable to think clearly'' (SNOMED ID 247640008), while the ground truth concept is ``Mentally dull'' (SNOMED ID 419723007). Finally, another group of errors is related to the problem of ambiguity of one-word expressions. For example,  ``stiff'' is automatically associated with ``Stiff legs'' (SNOMED ID 225609009) instead of ``Stiffness'' (SNOMED ID 271587009). We believe that these errors may be solved with advanced language modeling techniques, and suggest these models as future work.

\section{Related Work}
While there has been a lot of work on named entity recognition from UGTs and specifically social media posts done over the past 7 years~\citep{Leaman:2010:TIP:1869961.1869976,nikfarjam2015pharmacovigilance,karimi2015cadec,oronoz2015creation,SNG16,miftahutdinov2017identifying,vandam2017detecting}, relatively few researchers have looked at assigning social media phrases to medical identifiers. The first \emph{Social Media Mining shared task workshop} was designed to mine pharmacological and medical information from social media, with a competition based on a published dataset~\citep{SNG16}. Task~3 of this competition was devoted to medical concept normalization, where participants were required to identify the UMLS concept for a given adverse drug reaction (ADR). The evaluation set consisted of 476 ADR instances. \cite{SNG16} noted that there had been no prior work on normalization of concepts expressed in social media texts, and Task~3 did not attract much attention from the academic community.

Recently, two teams, namely UKNLP~\citep{Han201749} and GnTeam~\citep{Belousov201754}, participated in the \emph{Second Social Media Mining for Health} (SMM4H) \emph{Shared Task} and submitted their systems for automatic normalization of ADR mentions to MedDRA concepts. For Task~3,~\cite{sarker1overview} created a new dataset of phrases from tweets. The training set for this task contains $6{,}650$ phrases mapped to $472$ concepts, while the test set consisted of $2{,}500$ phrases mapped to $254$ classes. 
The systems presented by the teams showed similar results. The approach of \emph{GnTeam} contained three components for preprocessing and classification. The first two components corrected spelling mistakes and converted sentences into vector-space representation, respectively. For the third step, \emph{GnTeam} adopted multinomial logistic regression model which achieved the accuracy of $0.877$, while the bidirectional GRU achieved the accuracy of $0.855$. 
The ensemble of both classifiers showed slightly better performance and achieved the accuracy of $0.885$. UKNLP's system adopted a hierarchical LSTM in which a phrase is segmented into words and each word is segmented into characters. 
Hierarchical Char-LSTM achieved the accuracy of $0.872$, while hierarchical Char-CNN fared slightly better and achieved the accuracy of $0.877$. 

\cite{limsopatham2016normalising} experimented with CNNs and pre-trained word embeddings for mapping social media texts to medical concepts. For evaluation, three different datasets were used. The authors created two datasets with $201$ and $1{,}436$ Twitter phrases which mapped to concepts from a SIDER database. The third dataset is the CSIRO Adverse Drug Event Corpus (CADEC) \citep{karimi2015cadec} which consists of user reviews from \href{www.askapatient.com}{askapatient.com}. 
Experiments showed that CNN (accuracy $81$\%) outperformed DNorm (accuracy $73$\%), RNN (accuracy $80$\%) and multiclass logistic regression (accuracy $77$\%) on the \emph{AskAPatient} corpus (as well as corpora of tweets). This work is nearest to ours in the use of deep learning and semantic representation of words. However, we found that only approximately $40$\% of expressions in the test data are unique, while the rest of expressions occur in the training data. Therefore, the presented accuracy may be too optimistic. We believe that future research should focus on developing extrinsic test sets for medical concept normalization.


\section{Conclusion}
In this work, we have applied deep neural networks, in particular recurrent architectures based on LSTM and GRU units with attention, to the medical concept normalization problem for user-generated texts expressed in the free-form language of social networks. We have obtained very promising results, both quantitatively and qualitatively. As another contribution, we have added similarity features to the network and shown that this addition does further improve performance. 
We outline three directions for future work. First, novel recurrent architectures still look promising, and we believe there is still much to be explored there. Second, a promising research direction is to try and integrate linguistic knowledge into the models. Finally, a third direction of further research might focus on developing extrinsic test sets for concept normalization, a task that is sorely lacking now.

\textbf{Acknowledgements}\space This work was supported by the Russian Science Foundation grant no. 18-11-00284.

\bibliographystyle{acl_natbib}
\bibliography{ml}

\end{document}